\documentclass[11pt]{article}    
\usepackage[letterpaper,margin=1in]{geometry}

\usepackage{cite} 
\usepackage{url}  
\usepackage{ifthen}  
\usepackage{multicol}   
\usepackage[utf8]{inputenc} 
\usepackage{authblk}
\usepackage{graphicx}
\usepackage{amsmath,amsfonts,amssymb}
\usepackage{epsfig}
\usepackage{enumerate,enumitem}
\usepackage{multirow}
\usepackage{algorithm,algorithmic}
\usepackage{bm}
\usepackage{hhline}
\usepackage{array}
\newcolumntype{k}[1]{%
>{\raggedleft\hspace{0pt}}p{#1}}%
\newcolumntype{x}[1]{%
>{\centering\hspace{0pt}}p{#1}}%
\newcommand{\eps}{\epsilon}
\newcommand{\beps}{{\bm{\epsilon}}}
\newcommand{\R}{{\mathbb{R}}}
\newcommand{\N}{{\mathbb{N}}}

\newcommand{\eq}[1]{(\ref{#1})}

\begin{document}

\title{Inferring clonal evolution of tumors from single nucleotide somatic mutations}
 
\author[1,2]{Wei Jiao\thanks{Wei Jiao and Shankar Vembu contributed equally to this work.}}
\author[3]{Shankar Vembu}
\author[4]{Amit G. Deshwar}
\author[1,2]{Lincoln Stein}
\author[1,3,4,5,6]{Quaid Morris}
\affil[1]{Department of Molecular Genetics, University of Toronto}
\affil[2]{Ontario Institute for Cancer Research}
\affil[3]{Donnelly Center for Cellular and Biomolecular Research, University of Toronto}
\affil[4]{Edward S. Rogers Sr. Department of Electrical and Computer Engineering, University of Toronto}
\affil[5]{Banting and Best Department of Medical Research, University of Toronto}
\affil[6]{Department of Computer Science, University of Toronto}
\date{}

\maketitle

\begin{abstract}      
High-throughput sequencing allows the detection and
quantification of frequencies of somatic single nucleotide variants
(SNV) in heterogeneous tumor cell populations. In some cases, the
evolutionary history and population frequency of the subclonal
lineages of tumor cells
present in the sample can be reconstructed from these SNV
frequency measurements. However, automated methods to do this reconstruction are
not available and the conditions under which reconstruction is
possible have not been described.

We describe the conditions under which the
evolutionary history can be uniquely reconstructed from SNV
frequencies from single or multiple samples from the tumor population
and we introduce a new statistical model, \emph{PhyloSub}, that infers the phylogeny and genotype of the major subclonal
lineages represented in the population of cancer cells. It uses a Bayesian nonparametric prior over trees
that groups SNVs into major subclonal lineages and automatically estimates the number of lineages and
their ancestry. We
sample from the joint posterior distribution over trees to identify
evolutionary histories and cell population frequencies that have the
highest probability of generating the observed SNV frequency data. When multiple phylogenies are consistent with a given set
of SNV frequencies, PhyloSub represents the
uncertainty in the tumor phylogeny using a ``partial order plot.'' Experiments on a simulated dataset and two real datasets comprising tumor samples from acute myeloid leukemia and chronic lymphocytic leukemia
patients demonstrate that PhyloSub can infer both linear (or chain) and branching lineages and its inferences are in good agreement with ground truth, where it is available. PhyloSub can be applied to frequencies of any ``binary'' somatic mutation, including SNVs as well as small insertions and deletions. The PhyloSub and partial order plot software, and supplementary material are available from \hbox{\url{http://morrislab.med.utoronto.ca/phylosub/}}.

\end{abstract}

\section*{Background}
Cancer is a complex disease often associated with a characteristic series of somatic genetic variants \cite{Hanahan00,Hanahan11}.
Substantial effort has been devoted to genetic profiling of tumors in hopes of identifying these driver mutations and studying how they drive tumor development and resistance to treatment\cite{Yap12}. Tumors often contain multiple, genetically diverse subclonal populations of cells\cite{Visvader11}, and in some cases it is possible to reconstruct the evolutionary history of the tumor, thereby aiding in the identification of driver mutations, by computing the population frequencies of mutations that distinguish the subclonal populations \cite{Navin10,Mullighan08,Gerlinger12,Marusyk10,Schuh12,Shah12,Carter13,Landau13,OesperMR13}.

Somatic mutations can be detected, and roughly quantified, using exome and whole genome sequencing of a sample from a bulk tumor\cite{Marusyk12}.
However, recent attempts to reconstruct subclonal phylogenies have employed much deeper targeted sequencing \cite{Meyerson10} of tumor-associated single nucleotide variants (SNVs) to achieve higher accuracy in estimated SNV frequencies\cite{Shah12,nik12,Jan12,Schuh12}.
These SNV frequencies were then used to partially reconstruct the evolutionary history of tumors based on a single\cite{Shah12,nik12} or multiple \cite{Schuh12} samples of same tumor. However, due to short read sequencing, the frequencies of different SNVs are measured independently, so linkage between the SNVs in subclones is unavailable and standard phylogenetic methodology can not be used to construct evolutionary histories (as done in \cite{Campbell08} or \cite{Jan12}). However, if one makes the {\it infinite sites assumption} about tumor evolution, namely that every SNV only appeared once, then it is possible to use SNV frequencies to automatically reconstruct full or partial subclonal phylogenies while also inferring the multiple SNV genotypes of the major subclonal lineages in the tumor.

Here we describe a new method that automatically performs this phylogenetic reconstruction. First, we demonstrate that an unambiguous reconstruction is possible by describing {\it topological constraint rules} that are sufficient conditions to infer whether a triplet of SNV frequencies is consistent with only a chain or a branching phylogeny. We then describe a new method, {\it PhyloSub}, that automatically infers tumor phylogenies from SNV allele frequencies measured in single or multiple tumor samples. PhyloSub is based on a generative probabilistic model, inference in which implicitly implements the two rules by inferring the hidden phylogenies that have high probabilities of generating the observed SNV frequencies. It uses Bayesian inference, based on Markov Chain Monte Carlo (MCMC) sampling, to infer a distribution over phylogenies that incorporates uncertainty due to multiple phylogenies being consistent with the SNV frequencies and also noise in the measurement of the SNV frequencies. PhyloSub uses a Dirichlet process prior over phylogenies \cite{AdamsGJ10} to group SNVs into major subclonal lineages.

\subsection*{Model assumptions}
\begin{figure}[!t]
\centering
\includegraphics[scale=1]{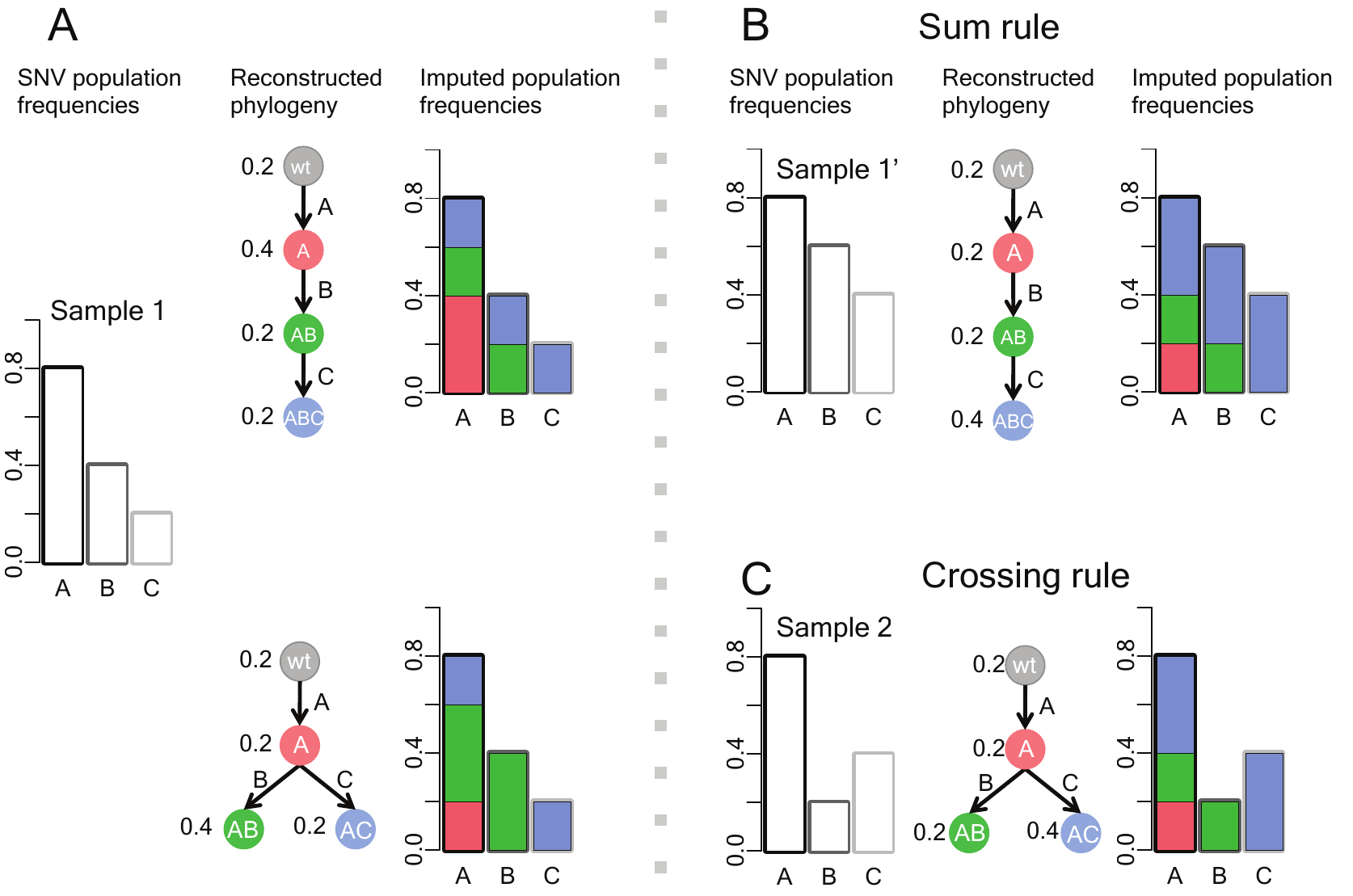}
\caption{{\bf Visualization of topological constraint rules.} {\bf (A)} A, B, C are three SNVs, each of which represents a set of SNVs with similar SNV population frequencies. When the SNV population frequencies are {0.8, 0.4, 0.2} (left panel), there might be two possible phylogenies that are consistent with these frequencies (middle panel). The two solutions estimate same number of clonal populations but different genotypes for each clone. The decomposition of the clonal population frequencies are shown on the right panel.
{\bf (B)} Because of the sum rule, for this given set of SNV population frequencies {0.8, 0.6, 0.4}, a chain structure may be the only possible phylogeny to explain the frequency changes.
{\bf (C)} Under the crossing rule, when multiple samples from the same patient are taken, we would expect the phylogenies are shared between samples. When another set of frequencies are observed {0.8,0.2,0.4}, the branching structure is the only possible phylogeny to explain the frequencies changes for both sample 1 and 2.
}
\label{fig:toy_example}
\end{figure}

We assume that the tumor evolution proceeds according to the clonal evolution theory, namely that all tumor cells are derived from ancestors that gain growth advantages over normal tissue and begins to expand \cite{Campbell08}. Subsequent mutations can provide a further fitness or survival advantage to their subclonal lineage\cite{Brosnan12} which subsequently increases in frequency compared to cells containing only the SNVs in the parental lineage. A given tumor sample is a snapshot of this evolutionary process and may contain, at non-negligible frequency, cells from multiple major subclonal lineages, each containing a different assortment of these advantageous mutations.
We make the {\it infinite sites assumption} \cite{kimura69,hudson83}, namely that each SNV appears only once and furthermore that once it appears, it does not revert back to its original state. As we describe below and illustrate in Figure \ref{fig:toy_example}, in some circumstances, this assumption highly constrains the phylogenies that are consistent with the SNV allele frequency data, especially if SNV frequencies from multiple samples from the same tumor are available. Finally, to make our model robust to low tumor cellularity, we assume that each tumor is derived from a single clone, however, this assumption is not critical in modeling tumor evolution and we revisit this assumption in the discussion section where we describe how to generalize our model to multicentral tumors (e.g., \cite{Shattuck05}). 

To simplify our initial discussion, we will assume that the exact population frequencies of the cells containing each SNV (i.e., the SNV population frequency) are available before discussing how we estimate these frequencies from deep sequencing data of the SNV locus. Note, we assume that the copy number of a locus is available as per \cite{Shah12}. In the datasets that we considered, most SNVs are heterozygous at a normal copy number locus and the population frequency of other SNVs is easily inferred from their allele frequencies. In more complex situations, a number of tools are available to infer copy number changes associated with specific subclonal lineages from whole genome sequencing data\cite{Carter13,OesperMR13}.

An important consequence of the infinite sites assumption is that if SNV B occurred in a cell that contained SNV A, then all cells that have B also have A and thus the population frequency of A must always be greater than or equal to that of B, regardless of where and when the tumor sample was taken.
However, a given set of three SNV population frequencies can still be consistent with two different phylogenies: a linear phylogeny or a branching phylogeny (see Figure \ref{fig:toy_example}A).

\subsection*{Topological constraint rules}

One can distinguish linear or branching descent under some circumstances. For example, if we have already established that SNV A is ancestral to both B and C (i.e., that all cells with B or C also contain A), then if the population frequency of B plus the population frequency of C is greater than the population frequency of A, then the phylogeny must be linear. This is true because in a branching phylogeny, there are no cells that contain both B and C, so the population frequency of A must be at least as large as sum of the frequencies of B and C (see Figure \ref{fig:toy_example}B). We call this the ``sum rule''. However, because a linear phylogeny is consistent with any set of SNV frequencies from a single sample, without making any further assumptions about the tumor evolution process, one needs at least two tumor samples to be able to rule out a linear phylogeny. However, given two samples and again assuming that SNV A is ancestral to both B and C, if the population frequency of B is larger than that of C in one sample, and vice versa in the other, than neither B nor C can be ancestral to the other, and the only phylogeny consistent with both sets of SNV frequencies is the branching one. We call this the ``crossing rule'' because the frequencies of B and C cross (see Figure \ref{fig:toy_example}C for an example). However, there is no guarantee that one can apply either rule to any set of SNV frequencies for all triplets of SNVs, although increasing the number of tumor samples does make it more likely that either the sum or crossing rule will be applicable for one or a pair of tumor samples, respectively. Furthermore, one needs to also consider the possibility of estimation error in the SNV population frequencies because these are inferred from discrete read counts. Note that these two rules also apply where SNV A is a mock SNV representing the wildtype state and having population frequency of $100\%$; as such these two rules also apply for multicentral tumors.

\subsection*{The PhyloSub algorithm}
\begin{figure}[!t]
\centering
\includegraphics[scale=.75]{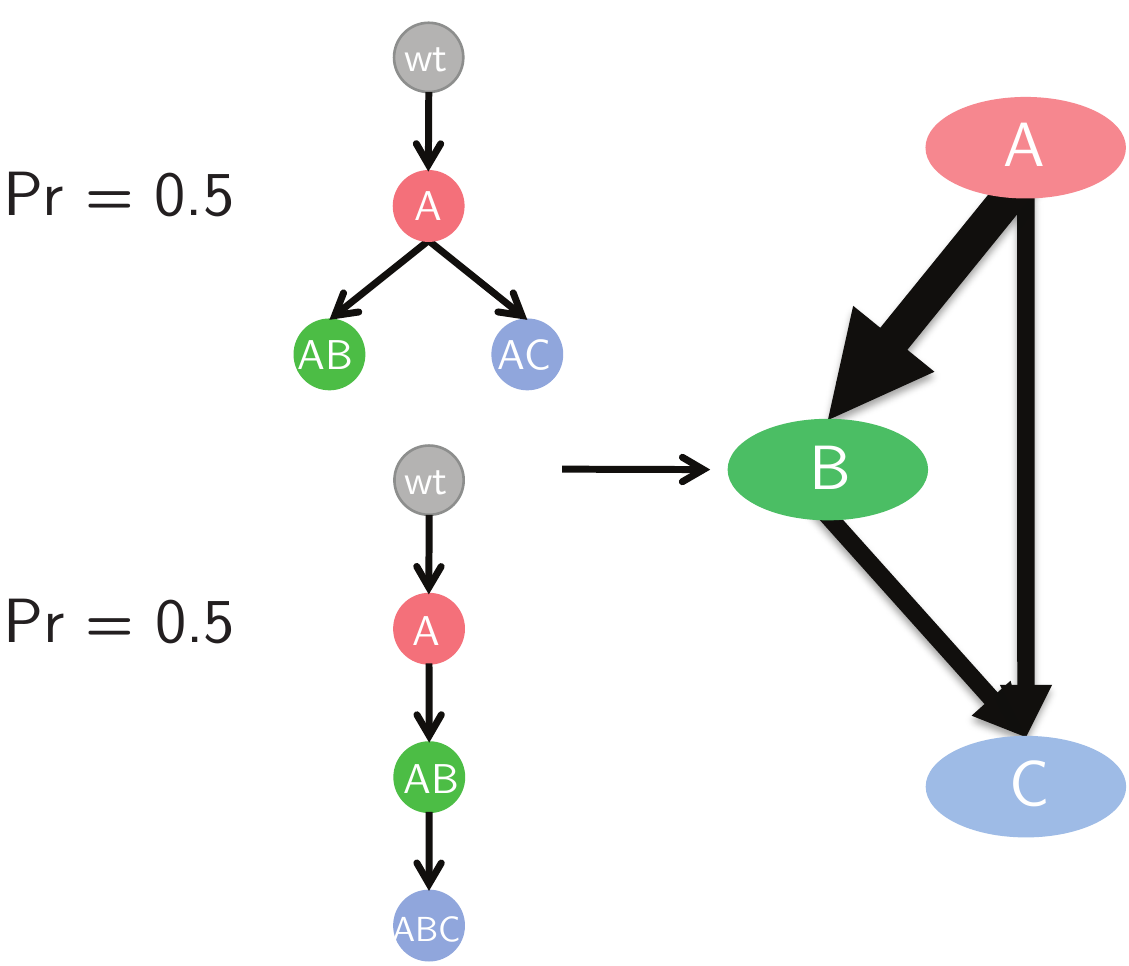}
\caption{{\bf Derivation of partial order plot.} (Left column) Posterior distribution over trees, each tree has a 0.5 probability under the posterior. (Right column) Derived partial order plot. Nodes correspond to SNVs. Edge thickness is proportional to the posterior probability that parent SNVs is in a subclonal lineage that is the parent of the one containing the child SNV. SNV nodes are ordered based on a topological sort implemented as the ``layered graph drawing method'' in Graphviz \cite{graphviz}. In this example, SNV A is always in a subclonal lineage that is the parent of B but the same is true with probability $0.5$ for C.}
\label{fig:toy_example2}
\end{figure}

To explicitly model uncertainty in estimates of the SNV population frequencies and the precise tumor phylogeny, we have developed the PhyloSub model that we describe here. PhyloSub attempts to explain the observed read counts in terms of a latent phylogeny that associates SNVs with particular subclonal lineages. PhyloSub takes as input a set of read counts for a set of SNVs and the copy number status of each SNV, and estimates the number of major subclonal lineages, the mutational profile of each lineage, and the proportion of each lineage within the tumor cell population from which the read data was drawn.
PhyloSub implements the parsimony assumptions detailed above using a non-parametric prior over tree structures. 
It is ``generative'' in that it attempts to explain the observed SNV frequencies in terms of an unobserved phylogeny; our model is also ``Bayesian'' in that it infers a posterior distribution over phylogenies and associated subclonal lineage frequencies.  We introduce a new visualization, the {\it partial order plot}, to represent the posterior uncertainty in the phylogeny when the SNV frequencies alone do not provide sufficient information to uniquely reconstruct the phylogeny (Figure \ref{fig:toy_example2}).  The sum and crossing rule described above are implicitly incorporated into our generative model -- our model assigns very low probability to any read counts that reflect deviations from either rule.

In the following, we provide a brief introduction to the PhyloSub model (see \textbf{Methods} for the full model) and we demonstrate its application to datasets where a single sample is profiled \cite{Jan12} and those where multiple samples are profiled \cite{Schuh12}. We also report the application of the model on a simulated dataset to show that its prior parameterization allows it to represent a wide variety of phylogenies.

\section*{Results and Discussion}
\subsection*{PhyloSub}
PhyloSub represents the major subclonal lineages and their evolutionary relationships using a directed tree in which each node is associated with a subclonal lineage and the edges connect parental lineages to their direct child lineage. Each subclonal lineage is associated with a distinct subset of the SNVs input to the model, we call this subset the  genotype of the lineage. Each node is also associated with (i) a set of SNVs that are present in this lineage but not its parent lineage and (ii) the population frequency of cells with the lineage genotype (and with no other SNVs from the input set). A subclonal lineage contains all of the SNVs associated with its parent, so its full genotype can be reconstructed by taking the union of the SNVs associated with its node and all of its ancestral nodes.  Similarly, the population frequency of an SNV is the sum of the subclonal lineage frequencies of the lineage it appeared in and all of its descendent lineages.  So, the subclonal lineage tree can be used to compute the population frequencies of each SNV and the genotype of each subclonal lineage. Associated with each SNV is a variable that indicates its zygosity and copy number in the cells that it appears (e.g., Aa indicating heterozygous and normal copy number), we assume all cells with the SNV have the same zygosity and copy number, and that all other cells have normal copy number at the SNV locus. The SNV genotype variable along with the population frequency is used to compute the allelic frequency of the SNV $i$, $p_i$. The data input to the model for each SNV is the number of reads mapping to the SNV locus, $d_i$, and the number of these reads that do not contain the SNV, $a_i$. We evaluate the likelihood of a given subclonal lineage tree (including the lineage population frequencies and the SNV genotype variables) by taking the product of the read count probabilities for each SNV, where the probability for the locus of SNV $i$ is computed using a binomial distribution whose parameter is derived from $p_i$ and an estimate of the error rate of the sequencer. PhyloSub also contains a vague prior over tree structures that is parameterized by three hyperparameters $(\alpha_0,\gamma,\lambda)$ (see \textbf{Methods}) that govern how the prior scores trees with more or fewer nodes, and different average numbers of siblings. We use ranges for these hyperparameters that in simulations have a slight preference for trees with fewer nodes but a limited preference for sibling numbers (see below for details).

\subsection*{Simulations}
PhyloSub's Dirichlet process prior over tree structures depends on three hyperparameters: $\alpha_0$, $\gamma$, and $\lambda$. The hyperparameters $\alpha_0$ and $\lambda$ determine the number of nodes (subclones) in the tree, $\lambda$ also affects the height of the tree and $\gamma$ affects the number of siblings in the tree which in turn affects the width of the tree. In all the experiments, we sample these hyperparameters \cite{AdamsGJ10} as part of the MCMC sampling from a range whose upper and lower bounds we establish in this section.

\begin{figure}[!t]
\centering
\includegraphics{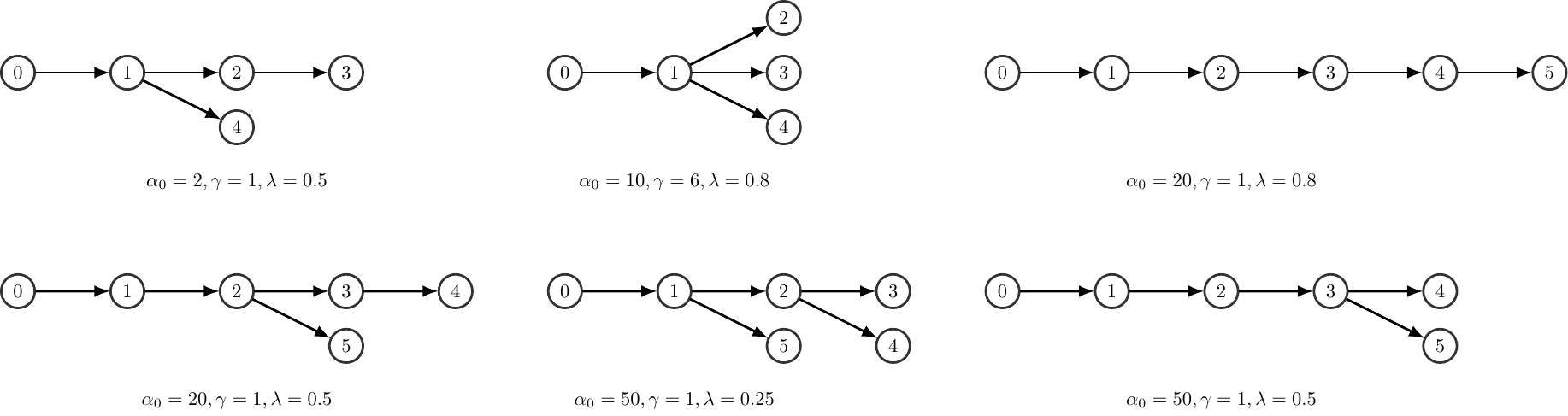}
\caption{{\bf Results on simulated dataset}. Best tree structures, i.e., the ones with the highest complete-data likelihood, estimated by PhyloSub with varying hyperparameters for the simulated dataset. We only show a subset of the trees from 90 MCMC runs corresponding to all possible combinations of the hyperparameters used in the simulation.}
\label{fig:tree_simulations}
\end{figure}

To establish the ranges that we use for the hyperparameters in PhyloSub, we simulated read counts from clusters with an average of nine SNVs per cluster with SNV population frequencies $\{1.0, 0.85, 0.6, 0.35, 0.2, 0.08\}$, with a read depth of $\approx 10,000$X which is a typical read depth for the targeting deep sequencing data that PhyloSub is designed for. We simulated heterozygous SNVs at loci with normal copy number and sample read counts for each SNV from a Binomial distribution (see \textbf{Methods}). The hyperparameter settings we used in the simulations are all possible combinations of $\alpha_0 \in \{1,2,4,10,20,50\}$, $\gamma \in \{1,2,4,6,8\}$ and $\lambda \in \{0.25,0.5,1\}$. The SNV population frequencies are consistent with many different tree structures and Figure \ref{fig:tree_simulations} shows that the tree structures with highest complete-data likelihoods varies in the expected way for different settings of the tree prior hyperparameters. Although the preferred structure varies, the inferred SNV frequencies remain well-correlated with the baseline values (Pearson correlation $> 0.99$) for these hyperparameter ranges, so the prior is not over-regularizing the SNV frequencies for these settings. To allow a range of tree structures, we integrate over these ranges by placing a uniform prior on the choice of these settings in our MCMC simulations (c.f., \cite{AdamsGJ10}). 

Although we focused on high read depths in the above simulation, we found that PhyloSub works well for read depths $\approx 1,000$X and was able to recover the clusters similar to the ones reported above and the SNV frequencies are well-correlated with the baseline values (Pearson correlation $> 0.99$). However, we found that the performance of the model degrades slightly at a read depth of $\approx 200$X, due to merging of clusters whose nearby SNV frequencies could not be distinguished. Nonetheless, we note that the inferred SNV population frequency estimates remain well-correlated (Pearson correlation $> 0.96$) and that the majority of the clusters were recovered at read depth $\approx 200$X.

The simulation as described above has no clear phylogeny by design. The SNV frequencies were consistent with multiple phylogenies and the main goal of this simulation was to establish the ranges for our hyperparameters that permit a wide variety of tree structures. We integrate over these parameter ranges on the real data in order to remove any prior bias towards particular structures. To determine whether PhyloSub can correctly recover the phylogenies from a single sample of SNV frequencies, we simulated read data from a chain phylogeny with SNV population frequencies $0.9 \to 0.75 \to 0.55 \to 0.4 \to 0.25$. By the sum rule, these frequencies are only consistent with a chain phylogeny. PhyloSub was able to reliably recover this chain. The real datasets described in the later sections are representative of the types of problems that our methodology could be applied to as they contain single and multiple samples, some of which have clear phylogenies and some do not.

\subsection*{Results on AML datasets}
To assess PhyloSub on single samples, we applied it to data from Jan \emph{et al.} \cite{Jan12} who reported the coexistence of multiple subclonal lineages in hematopoietic stem cells (HSC) from acute myeloid leukemia (AML) patient samples. The deep-targeted sequencing of all SNV candidates identified by exome sequencing identified SNVs with differing allelic frequency, suggesting multiple clonal populations in the HSC cells. An independent single-cell assay confirmed the existence of multiple clones, and thus provides a ground truth tree that shows some of the major subclonal lineages within the populations. Here we apply PhyloSub to the two samples profiled by Jan \emph{et al.} that had three or more SNVs profiled in a single-cell assay.  These samples are SU048 and SU070 which have 6 and 10 SNVs in the single-cell assay, respectively. Although this assay confirmed the presence of some of the subclonal lineages, only 100-200 cells were assayed, so lineages with low population frequency in the sample (e.g., $< 1\%$) may not be detected.

We applied PhyloSub providing it with the copy number and zygosity of each SNV (results were similar if we assume normal copy number and have a uniform prior on zygosity).  For both SU048 and SU070, a number of different phylogenies were consistent with the SNV read counts, and we developed the ``partial order plot'' to represent the posterior uncertainty in the phylogeny (see Figure \ref{fig:toy_example2} and \textbf{Methods}).

\begin{figure}[!t]
\centering
\includegraphics[scale=1]{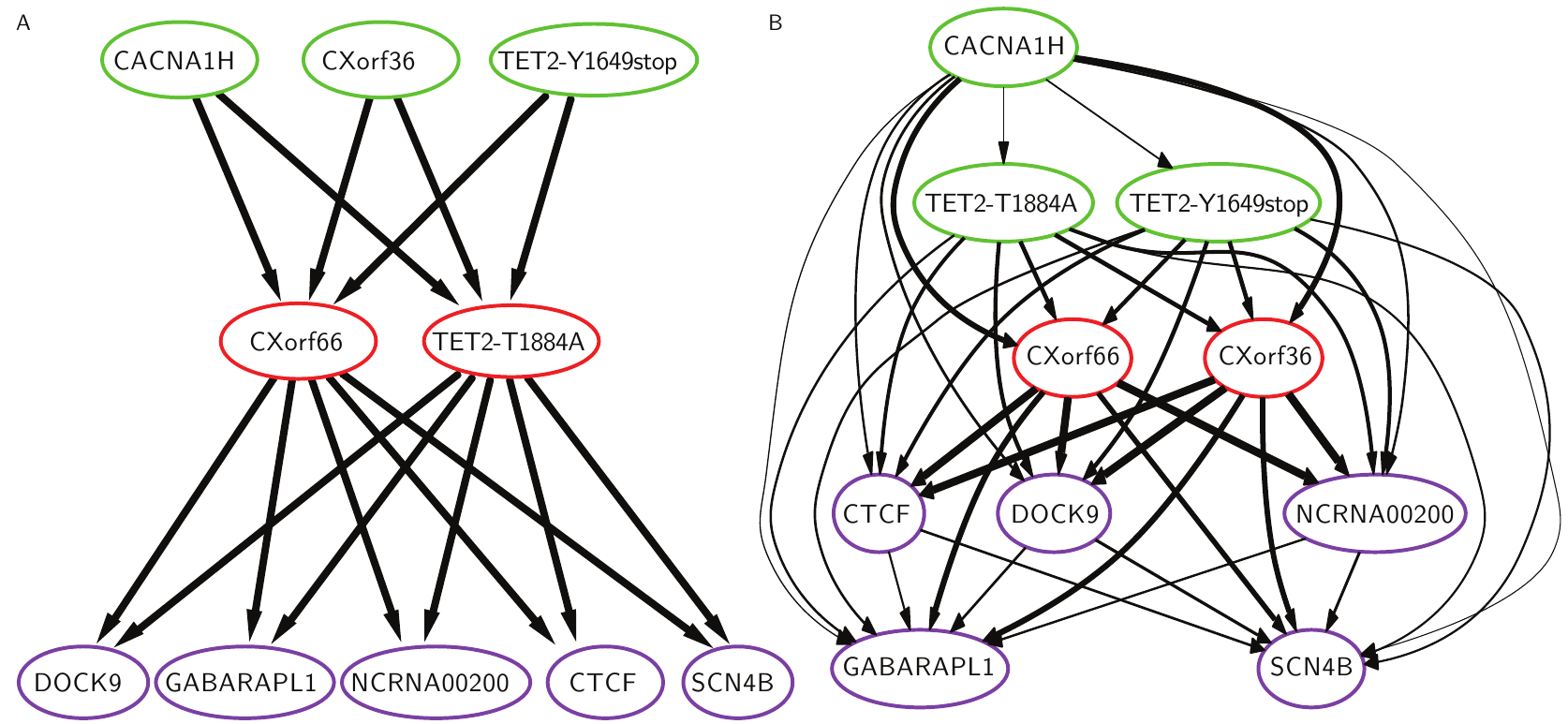}
\caption{{\bf Clonal evolutionary structures of tumor sample SU070.}  {\bf (A)} Ground truth from  Jan \emph{et al.} \cite{Jan12}.  {\bf (B)} PhyloSub's output summarized using a partial order plot. 
For clarity, we removed edges with probability less than $0.1$ before laying out the nodes and we only show SNVs for which single-cell data is available. However, all the SNVs whose frequencies were reported in this study were used in the inference. The color of the border of the SNVs represents the subclonal lineage cluster that the SNV is placed into by a graph-based clustering algorithm that takes as input the co-clustering frequencies from the MCMC samples (see \textbf{Methods}). Note that unlike the thickness of the edges, this is simply a visualization aid, and does not fully represent the model's posterior uncertainty in the SNV clusterings. 
}
\label{fig:tree_su070}
\end{figure}

Figure \ref{fig:tree_su070} shows that partial order plot for SU070. The ordering of the nodes in the partial order plot can also be used to infer ancestry via transitivity, for example, in Figure \ref{fig:tree_su070}, the SNV CXorf66 has high probability of being in the subclonal lineage that is the direct parent of the one that DOCK9 is in, however, because the TET2-T1884A SNV is sorted before CXorf66 (and has a small probability of being a direct parent of it), then in the PhyloSub posterior over lineages, TET2-T1884A has a high probability of being in an ancestral lineage to the one CXorf66 is in.  

Furthermore, one can interpret the partial order plot to indicate that both CXorf36 and CXorf66 are in the same lineage because they are both direct parents of DOCK9 (with high probability) and there are no edges between them. For reference, in Figure \ref{fig:tree_su070} we have included the results of the single-cell assay for SU070 in the partial order plot representation -- Jan \emph{et al.} report three subclonal lineages for SU070, as indicated by the SNV colorings\cite{Jan12}. We note that these plots are largely consistent. Indeed, we assign high posterior probability $>0.96$, to two of the three subclonal lineages detected by Jan \emph{et al.} (see Table S3 in Additional file 2 for full lineage genotype probabilities). For reference, we also provide the list of the subclonal lineage trees along with their posterior probabilities in Table S1 (see Additional file 2).

The one major difference between PhyloSub's estimates and the single-cell data from Jan \emph{et al.} is that PhyloSub switches the order of the appearance of SNVs CXorf36 and TET2-T1884A. In fact, there was not a single subclonal lineage that contained CXorf36 but not TET2-T1884A in 5,000 subclonal lineage trees sampled from PhyloSub's posterior. This switch is likely due to the observed SNV frequencies, indeed the 95\% confidence intervals of the SNV frequencies of these two SNVs do not overlap (Table \ref{tbl:app_su070}).
\begin{table}[!t]
\caption{Allelic counts for tumor sample SU070 from  Jan \emph{et al.} \cite{Jan12}.\label{tbl:app_su070}}
\begin{center}
\begin{tabular}{l|k{2.5cm}|r|x{4.75cm}|x{1.5cm}}
SNV	 & Variant allele read counts  & Read depth & Allele frequency & Cluster ID \tabularnewline
\hline

CACNA1H &	12,085	& 24,860 &	0.486 (95$\%$ CI: 0.481-0.491) &	A\tabularnewline
TET2-T1884A &	4,220 & 	8,772	& 0.481 (95$\%$ CI: 0.472-0.490) &	B\tabularnewline
TET2-Y1649stop	 & 7,792	 & 16,211	& 0.481 (95$\%$ CI: 0.474-0.487) &	A\tabularnewline
CXorf66  & 	3,684 &	8,150 &	0.452 (95$\%$ CI: 0.443-0.461)&B\tabularnewline
CXorf36 &	3,523	& 8,060 	&	0.437 (95$\%$ CI: 0.428-0.446)	&A\tabularnewline
DOCK9 &	3,391	& 8,676	& 	0.391 (95$\%$ CI: 0.382-0.400) &C\tabularnewline
NCRNA00200 &	9,201 &	25,413 &	0.362 (95$\%$ CI: 0.357-0.367)	&C\tabularnewline
CTCF	& 10,558 &	30,119 & 		0.351 (95$\%$ CI: 0.346-0.355)	& C\tabularnewline
GABARAPL1 	& 1,648 & 	4,992	 &	0.330 (95$\%$ CI: 0.319-0341) & C\tabularnewline
SCN4B &	5,113 &	16,386 &		0.312 (95$\%$ CI: 0.306-0.318) & C\tabularnewline

\end{tabular}
\end{center}
\end{table}
One explanation for this difference is a systematic bias in the measurement of one or both of these SNVs; it is also possible that the labels of these two SNVs were switched in Jan \emph{et al.}. We also note, however, that in Jan \emph{et al.}, the existence of the lineage that contains only CXorf36, TET2-Y1649stop, and CACNA1H is only supported by 2 of the 189 clones that they profiled.

\begin{figure}[!t]
\centering
\includegraphics[scale=1]{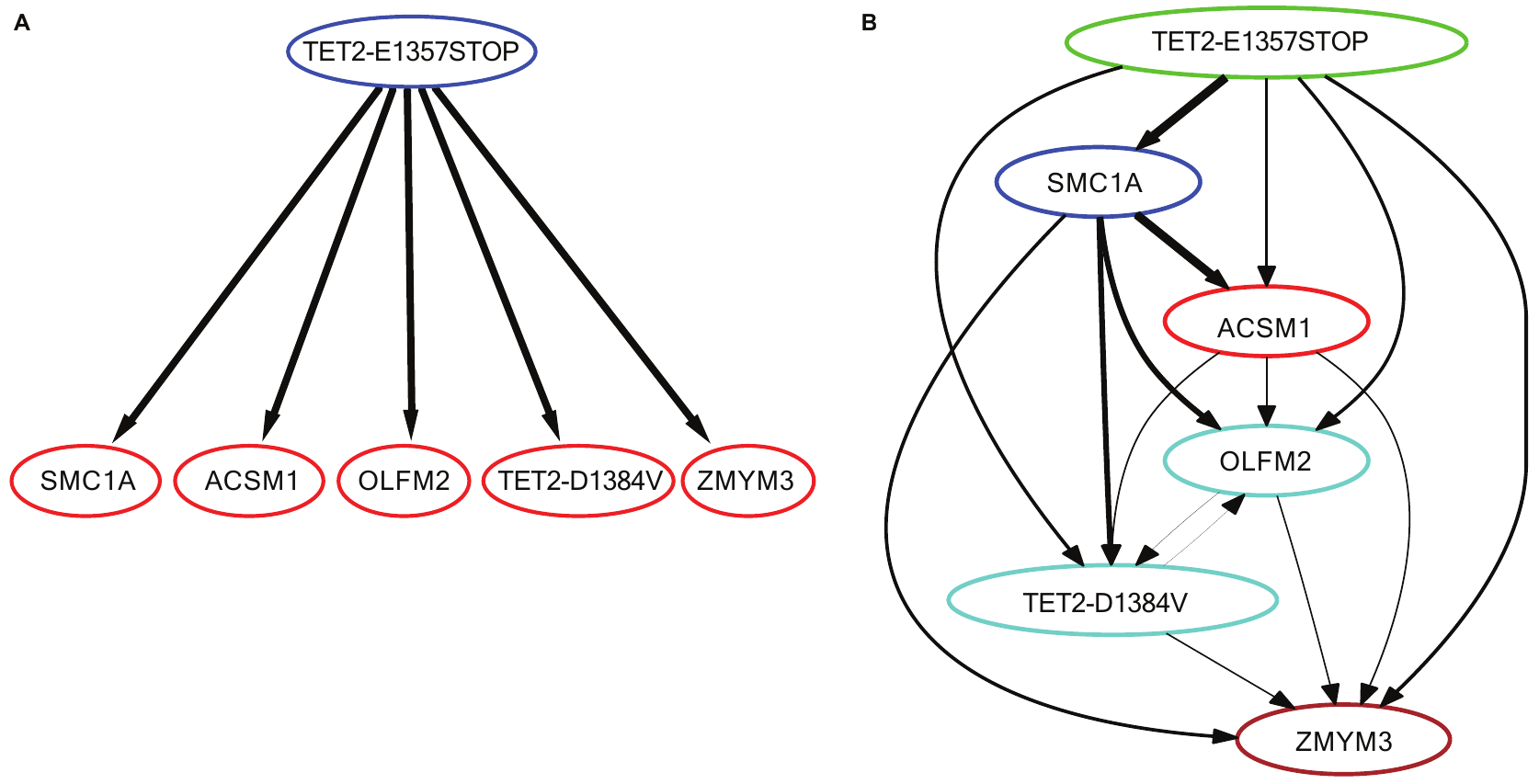}
\caption{{\bf Clonal evolutionary structures of tumor sample SU048.}   {\bf (A)} Ground truth from  Jan \emph{et al.} \cite{Jan12}.   {\bf (B)} PhyloSub's output summarized using a partial order plot. See Figure \ref{fig:tree_su070} legend for more details on partial order plot.}
\label{fig:tree_su048}
\end{figure}

For the tumor sample SU048, both the partial order plot and the single-cell assay agree on TET2-E1357stop event occurring early (at the root of the tree), and all other SNVs are secondary mutational events as shown in Figure \ref{fig:tree_su048}B. Note that the partial order plot shows a large uncertainty in the structure for the rest of the SNVs and this is also reflected in the posterior over subclonal lineage trees and genotypes (see Tables S2 and S4 in Additional file 2, respectively). There is no strong evidence for either a linear or branching lineage or for particular clustering among these SNVs. Also, from Table \ref{tbl:app_su048}, we see a lot of variation in the allele frequencies of these SNVs suggesting that they may not belong to the same subclonal lineages.
\begin{table}[!t]
\caption{Allelic counts for tumor sample SU048 from  Jan \emph{et al.} \cite{Jan12}. \label{tbl:app_su048}}
\begin{center}

\begin{tabular}{l|k{2.5cm}|r|x{4.75cm}|x{1.5cm}}
SNV	 & Variant allele read counts	 & Read depth	& Allele frequency & Cluster ID \tabularnewline
\hline
TET2-E1357stop &	7,436	& 19,553 &	0.380 (95$\%$ CI: 0.375-0.386) &	A\tabularnewline
SMC1A &	182,974 &	660,069	&0.277 (95$\%$ CI: 0.276-0.278) &	B\tabularnewline
ACSM1 &	17,149	 & 127,236	&	0.135 (95$\%$ CI: 0.133-0.136)&	B\tabularnewline
OLFM2 &	13,828 &	122,523	&	0.113 (95$\%$ CI: 0.111-0.114)&	B\tabularnewline
TET2-D1384V &	1,833	& 17,687 & 0.104 (95$\%$ CI: 0.100-0.107) &	B\tabularnewline
ZMYM3 &	18,536 &	307,346	& 0.060 (95$\%$ CI: 0.060-0.061)	&B\tabularnewline
\end{tabular}
\end{center}
\end{table} 
The subclonal lineage inferred by Jan \emph{et al.}'s single-cell assay is shown in Figure \ref{fig:tree_su048}A and only contains two lineages, one with only TET2-E1357stop and the other with the other five SNVs. The TET2-E1357stop lineage genotype has probability $0.81$  in our posterior, however the second genotype has a relatively small probability ($0.06$) under the posterior although we note that the genotype that contains all SNVs but ZMYM3 has a posterior probability of $0.32$ (see Table S4 in Additional file 2). For reference, we also provide the list of the subclonal lineage trees along with their posterior probabilities in Table S2 (see Additional file 2).

In summary, the subclonal lineage trees inferred by PhyloSub on single samples of SNV frequencies are largely consistent with ground truth but there remains substantial uncertainty in SU048 about whether there was a linear or chain lineage. On the other hand, the SNV frequencies in SU070 were only consistent with a linear lineage and PhyloSub almost perfectly reconstructed the results of the single-cell assay with one misordering of the SNVs. This difference may be explained by unmodeled systematic biases in the deep sequencing data or experimental error. Nonetheless, we have shown that in some cases, it is possible to achieve a good estimate of the genotype of multiple subclonal lineages as well as their evolution from a single, targeted deep sequencing sample of SNV frequencies.

\subsection*{Results on CLL datasets}
To evaluate PhyloSub on a multiple sample dataset, we used data from a study of chronic lymphocytic leukemia (CLL)  by Schuh \emph{et al.} \cite{Schuh12}  which quantified SNV frequencies of a set of SNVs during different time points spanning the patient therapy cycle. The candidate SNVs were identified by exome sequencing and then subjected to targeted resequencing. The tumor samples from the three patients in the study labeled CLL077, CLL006 and CLL003 have 11, 16 and 20 SNVs respectively with SNV frequencies for five different time points. Originally, Schuh \emph{et al.} reconstructed the evolutionary histories of each cancer by a semi-manual procedure in which they first automatically grouped SNVs into subclonal lineages using $k$-means clustering on the allele frequencies and the differences in allele frequencies between the time points for each patient and then reconstructed the evolutionary structure of those lineages using a procedure that they do not describe in the paper. In PhyloSub, we model multiple samples from the same cancer as sharing the same evolutionary history but we allow subclonal frequencies to change between samples.

We applied PhyloSub to the SNV read count data, providing the algorithm with the likely zygosity estimates -- in most cases, SNVs appeared to be heterozygous with normal copy number but in a few cases, SNVs appeared to be hemizygous and were input to the model as such. For these data, because of the multiple samples per tumor, there is very little posterior uncertainty in the best fitting tree -- as such, we only show the best single tree structure corresponding to the MCMC sample with the highest complete-data likelihood \cite{AdamsGJ10}.

\begin{figure}[!t]
\centering
\includegraphics[]{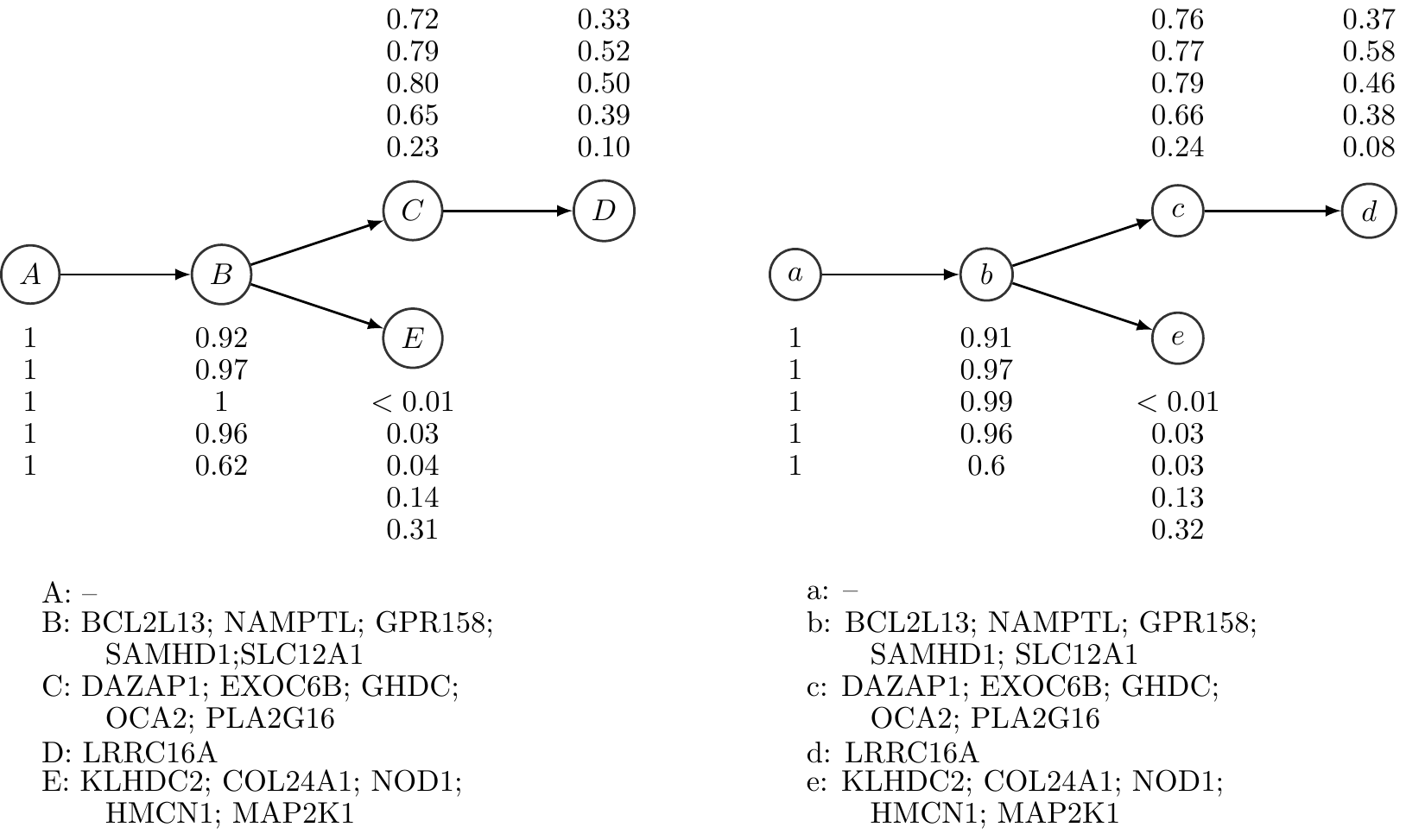}
\caption{{\bf Clonal evolutionary structures of tumor samples from patient CLL077.}   {\bf (Left)} Baseline tree structure from Schuh \emph{et al.} \cite{Schuh12}.  {\bf (Right)} Best tree structure estimated by PhyloSub. The SNV population frequencies and the cluster assignments are also shown in the figure.}
\label{fig:multi_cll077}
\end{figure}

\begin{figure}[!t]
\centering
\includegraphics[]{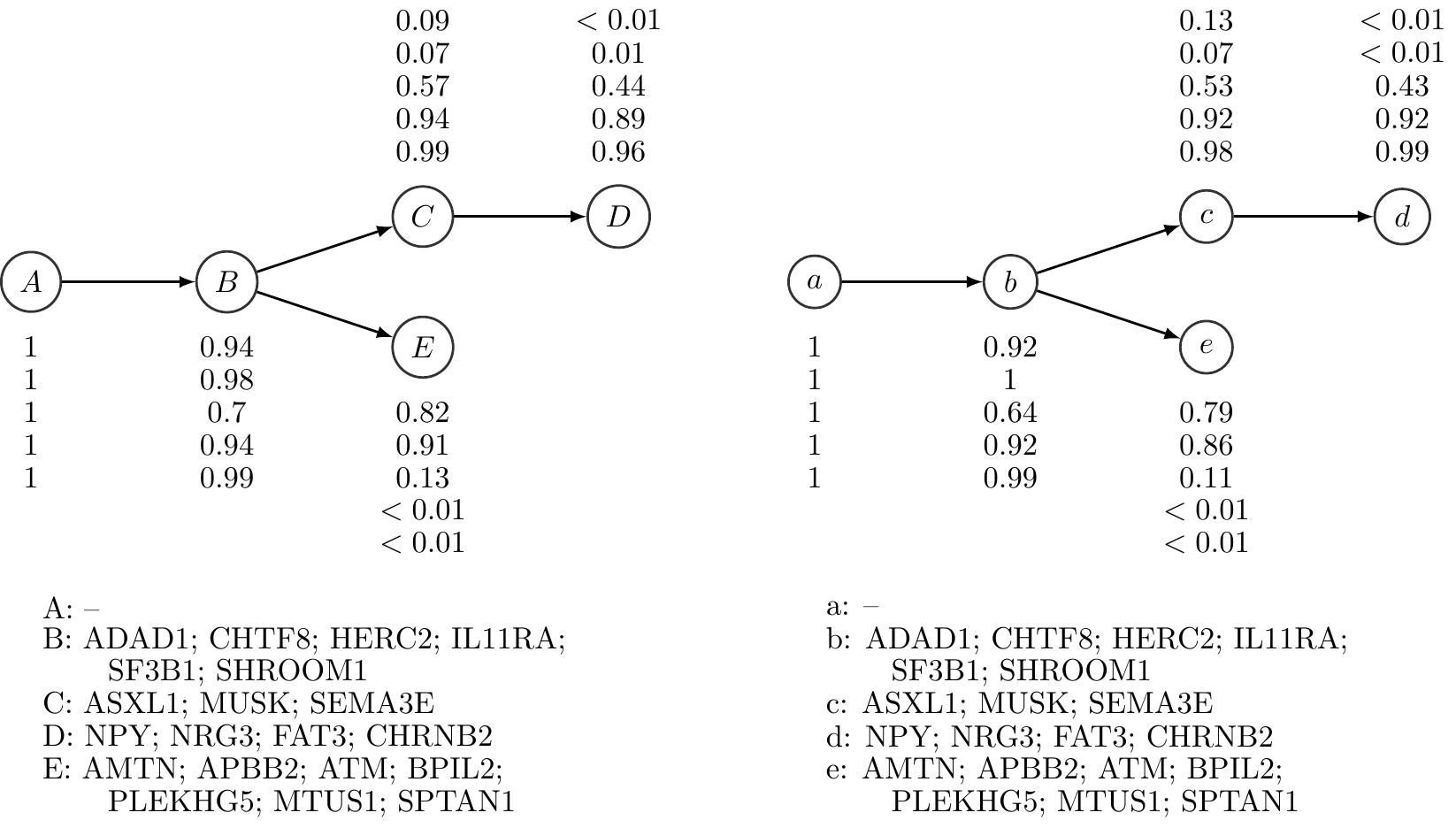}
\caption{{\bf Clonal evolutionary structures of tumor samples from patient CLL003.}  {\bf (Left)} Baseline tree structure from Schuh \emph{et al.} \cite{Schuh12}. {\bf (Right)} Best tree structure estimated by PhyloSub. The SNV population frequencies and the cluster assignments are also shown in the figure.}
\label{fig:multi_cll003}
\end{figure}

For the tumor samples CLL077 and CLL003, the best tree structure estimated by PhyloSub and the tree structure from Schuh \emph{et al.} \cite{Schuh12} are in exact agreement and the population frequencies of the subclonal lineages are well-correlated. Figures \ref{fig:multi_cll077} and \ref{fig:multi_cll003}  compare the PhyloSub estimates with those reported by Schuh \emph{et al.} \cite{Schuh12}. 

\begin{figure}[!t]
\centering
\includegraphics[]{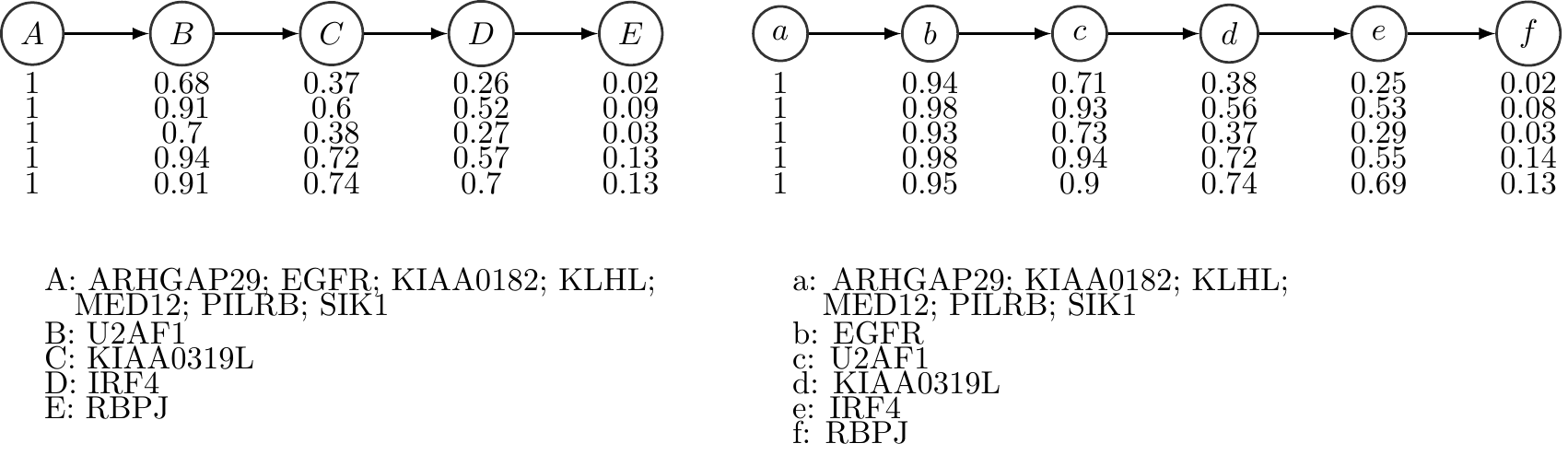}
\caption{{\bf Clonal evolutionary structures of tumor samples from patient CLL006.}   {\bf (Left)} Baseline tree structure from Schuh \emph{et al.} \cite{Schuh12}.  {\bf (Right)} Best tree structure estimated by PhyloSub. The SNV population frequencies and the cluster assignments are also shown in the figure.}
\label{fig:multi_cll006}
\end{figure}

For the tumor sample CLL006, PhyloSub inferred a chain structure similar to the chain structure from Schuh \emph{et al.}, but the major difference in PhyloSub's best estimate of the tree structure is the splitting of cluster A into two clusters as shown in Figure \ref{fig:multi_cll006}. However, we found that the complete-data log likelihood of PhyloSub's best estimate of the tree structure is higher than the one for the chain structure  of  Schuh \emph{et al.} and therefore PhyloSub prefers the splitting of the cluster A into two clusters.


\begin{figure}[!t]
\centering
\includegraphics[scale=1]{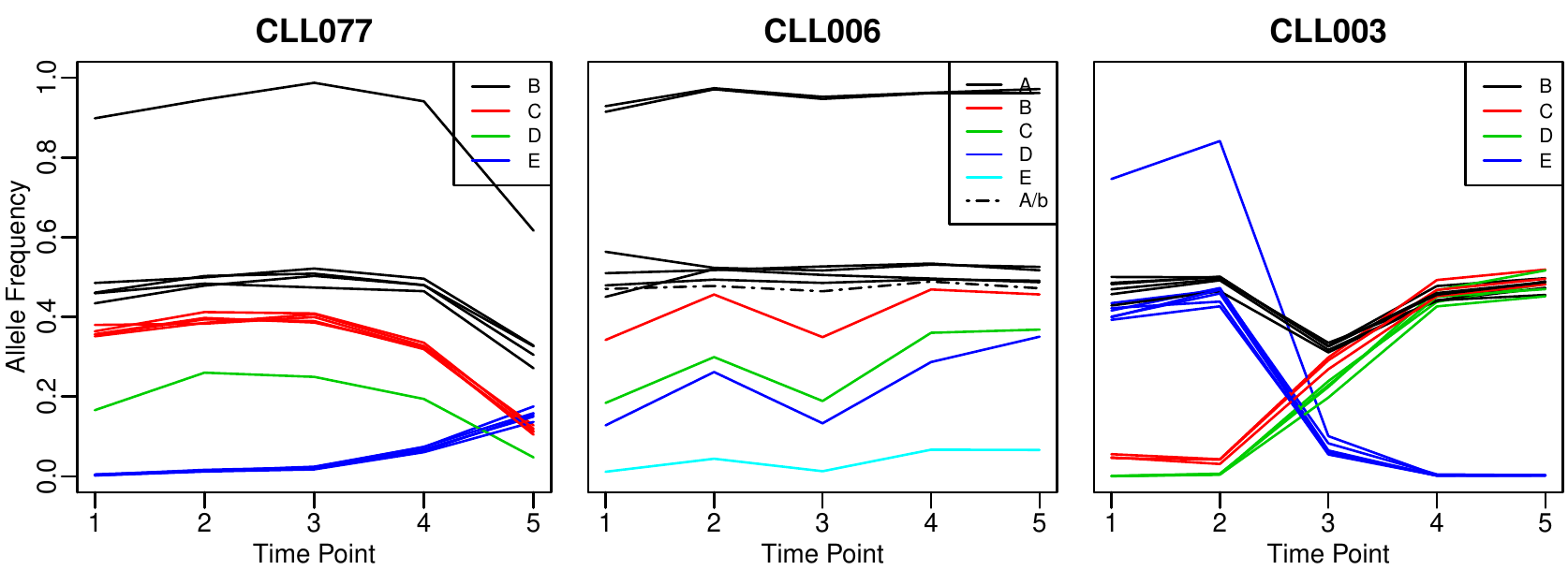}
\caption{{\bf Allele frequencies in the CLL datasets.} Changes in allele frequency with time point for the multiple tumor samples in CLL077, CLL006, CLL003 datasets from Schuh \emph{et al.} \cite{Schuh12}. Colors indicate SNV clusters. The dotted line in the middle panel plot indicates the SNV that formed its own cluster in PhyloSub's estimate of the tree structure.}
\label{fig:cll_frequency_plot}
\end{figure}

In the CLL dataset, there is no ground truth but to allow the reader to compare the two estimates of the evolutionary history, Figure \ref{fig:cll_frequency_plot} plots the frequency of each SNV in the three samples, and we have colored SNVs according to their subclonal lineage assignments by Schuh \emph{et al.}. These SNV frequencies are not corrected for copy number, however, the hemizygous SNVs are clear from examination of the figure. 

In summary, having multiple samples of SNV frequencies greatly reduces the posterior uncertainty in the evolutionary history of the tumor and PhyloSub is able to reconstruct histories produced by a semi-manual procedure.

\section*{Conclusions}
We presented a nonparametric Bayesian model called PhyloSub that uses
a Dirichlet process prior over trees \cite{AdamsGJ10} to model
the clonal evolutionary structure of tumors from next generation
sequencing data. We also introduced a new visualization method, the
partial order plot, to represent the posterior uncertainty in the
phylogeny when the clonal frequencies alone do not provide sufficient
information to uniquely reconstruct the phylogeny and mutational
profiles of each subclonal lineage represented in the tumor. By
enforcing a set of structural constraints on the SNV population frequencies
using MCMC methods, we were able to infer the phylogenetic
relationships between subclones from both single and multiple tumor
samples.

We have demonstrated that it is possible, in some cases, to detect a
linear lineage from a single, high cellularity sample of the tumor. We
have also shown that multiple samples highly constrain the possible
lineages that are consistent with the SNV frequency data.  
PhyloSub's inferred subclonal lineage trees  were in good agreement with single cell assays on single sample data and with an expert-driven, semi-manual reconstruction procedure on
multiple sample data.

PhyloSub’s ability to detect and characterize subclonal lineages depends on the frequency of the lineage in the population (compared to its descendant lineages), the number of SNVs that define the lineage, as well as the accuracy with which the SNV population frequencies are estimated which depends on both the sequencing depth as well as uncertainty about the copy number of the SNV. Simply put, for lineages defined by a single SNV, the read depth has to be high enough that the uncertainty in the estimated SNV frequency is less than the frequency of the subclonal population. Having more lineage-defining SNVs can relax this hard constraint. As such, the phylogenies of tumors with large numbers of subclonal lineages, each defined by a small number of SNVs (possibly due to a pronounced hypermutability phenotype), will be hard to reconstruct with PhyloSub, or any other method, unless the SNV frequencies are very accurately estimated. Indeed, it is not clear how ground truth could be uncovered in such a case: the gold standard of single cell sequencing would require an exceptionally large number of single cells to survey this highly heterogeneous population, and each of these cells would need to be sequenced deeply in order to ensure precise somatic variant calling.

One potential difficulty in scaling our approach to orders of magnitude more SNVs is that the Markov chain may not mix in a timely manner, in other words, may get stuck in local minima. We note that finding suboptimal solutions is an issue for any method based on these data. In our case, the mixing time of the chain would depend largely on the number of subclones represented in the population with less of a dependence on the number of SNVs. There are various techniques for determining whether or not a Markov chain is well-mixed and we refer the reader to a recent excellent review \cite{LevinPW2008}.

PhyloSub extends recent work on inferred cellularity and subclonal structure from somatic mutations. ABSOLUTE uses whole genome sequence data or array CGH data to identify regions of copy number change in the tumor and based on this infers cellularity and copy number changes associated with different subclones \cite{Carter13}. THetA \cite{OesperMR13} also attempts to infer both the copy number profiles and their relative proportions using the whole genome sequencing data based on an infinite sites assumptions. Neither of these algorithms explicitly reconstructs tumor phylogenies. Our work is closest to PyClone \cite{Shah12} which uses a flat Dirichlet process mixture model to group SNVs into subclonal lineages based on their frequencies; PhyloSub extends this work by reconstructing the phylogenetic relationships among these lineages and, in doing so, allows the full SNV genotype of each subclonal population to be reconstructed.

We designed PhyloSub to assume a single clonal origin for the cancerous
cells in the sample. We made this decision to increase the applicability of the sum rule for
low cellularity tumors (i.e., tumors with high normal contamination). However, removing this assumption would be a simple change to the model, which we have not evaluated.

Another area of future innovation would be in modeling sequencing
biases and uncertainty in SNV allele frequencies resulting from
them. We did not evaluate replacing our binomial model with a negative
binomial one that would have allowed greater variability in the
observed read counts for a given SNV allele frequency \cite{RobinsonO10}.

\section*{Methods}
\subsection*{Dirichlet process mixture models}
Consider the problem of clustering $N$ objects $\{x_i\}_{i=1}^N$ using a Bayesian finite mixture model of $K$ components (clusters) with the following generative process \cite{Teh2010}:
\begin{equation}\label{eqn:fmm}
\begin{aligned}
\bm{\omega}  \sim \text{Dirichlet}(\alpha/K,\ldots,\alpha/K)\ ; \quad
& z_{i}  \sim \text{Multinomial}(\bm{\omega})\ ; \quad
\phi_{k} \sim  H\ ; \quad
& x_{i}  \sim  F(\phi_{z_i}) \ ,
\end{aligned}
\end{equation}
where $\bm{\omega}$ are the mixing weights such that $\sum_{k=1}^K \omega_k=1$, $\alpha$ is the concentration parameter of the symmetric Dirichlet prior placed on the mixing weights, $z_i \in \{1,\ldots,K\}$ is the cluster assignment variable, $H$ is the prior distribution from which the component parameters $\{\phi_k\}$ are drawn, $F(\phi)$ is the component distribution parameterized by $\phi$. The finite mixture model can be extended to a model with an infinite number of mixture components by replacing the Dirichlet prior with a Dirichlet process (DP) prior resulting in what is known as the DP mixture model (DPMM) \cite{Antoniak74}.
Unlike finite mixture models, DPMMs automatically estimate the number of components from the data thereby  circumventing the problem of fixing the number of components \emph{a priori}. The stick-breaking construction \cite{Seth94} given below provides a precise recipe to draw samples from a Dirichlet process:
\begin{equation}
\label{eqn:stick}
\begin{aligned}
\beta_k \sim \text{Beta}(1,\alpha) \ ; \quad
\omega_1 = \beta_1 \ ; \quad
\omega_{k}  = \beta_k \prod_{\ell=1}^{k-1}(1-\beta_\ell) \ ; \quad
\phi_{k} \sim  H \ ; \quad
\mathcal{G} &= \sum\limits_{k=1}^\infty \omega_k \delta_{\phi_k} \ ,
\end{aligned}
\end{equation}
where $\delta_{\phi}$ is a point mass centered at $\phi$ and $\mathcal{G} \sim \text{DP}(\alpha,H)$, i.e., $\mathcal{G}$ is a draw from a DP with base distribution $H$ and concentration parameter $\alpha$. The stick-breaking process can be viewed as recursively breaking sticks of length $\prod_{\ell=1}^{k-1}(1-\beta_\ell)$, starting with a stick of unit length. The beta variates $\{\beta_k\}$ determine the random location at which the stick is broken. The concentration parameter $\alpha$ determines the number of clusters with high values resulting in large number of clusters. Let $\text{GEM}(\alpha)$ denote the stick-breaking process over $\bm{\omega}$. Replacing the Dirichlet prior in the finite mixture model \eq{eqn:fmm} with the stick-breaking process prior results in the following generative process for infinite mixture models:
\begin{equation*}\label{eqn:dpmm}
\begin{aligned}
\bm{\omega}  \sim \text{GEM}(\alpha) \ ; \quad
z_{i}   \sim \text{Multinomial}(\bm{\omega}) \ ; \quad
\phi_{k} \sim  H \ ; \quad
x_{i}   \sim  F(\phi_{z_i}) \ .
\end{aligned}
\end{equation*}
An alternative view of the above generative process  produces component parameters $\{\tilde{\phi}_i\}$ by drawing samples from $\mathcal{G}$ resulting in the following generative process:
\begin{equation}\label{eqn:dpmm1}
\begin{aligned}
\mathcal{G}  \sim \mathrm{DP}(\alpha,H) \ ; \quad
\tilde{\phi}_{i}  \sim \mathcal{G} \ ; \quad
x_{i}  \sim  F(\tilde{\phi}_{i}) \ .
\end{aligned}
\end{equation}
Note that in the above process every object $\{x_i\}_{i=1}^N$ is associated with a component parameter $\{\tilde{\phi}_i\}_{i=1}^N$ and that all objects assigned to the same cluster will have the same component parameter. In other words, multiple elements in the set $\{\tilde{\phi}_i\}_{i=1}^N$ will take on the same value from the set $\{\phi_k\}_{k=1}^K$ of unique parameters. 

\subsection*{Tree-structured stick-breaking process}
The stick-breaking construction \eq{eqn:stick} described above can be used to produce a \emph{flat} clustering of objects, where the clusters are independent of each other. Adams \emph{et al.} \cite{AdamsGJ10} extended this construction for hierarchical clustering by interleaving two stick-breaking processes. This construction results in a \emph{relational} clustering of objects where the clusters are connected to form a rooted tree structure. Unlike classical hierarchical clustering algorithms such as agglomerative clustering, this construction allows data to reside in the internal nodes of the tree; a feature we exploit to model the association of SNVs with subclonal lineages.

We borrow notation from Adams \emph{et al.} \cite{AdamsGJ10}. Let $\beps=(\epsilon_1,\ldots,\epsilon_p)$ denote a sequence of positive integers used to index the nodes of the tree. Let $\beps=\kappa$ denote the zero-length string, i.e., the root of the tree. Let $|\beps|$ indicate the length of the sequence $\beps$ and therefore the depth of node $\beps$. Let $\beps\eps_i$ denote the sequence formed by appending $\eps_i$ to $\beps$. The children of node $\beps$ is the set $\{\beps\eps_i : \eps_i \in 1,2,\ldots\}$ and let the ancestors of $\beps$ be denoted by the set $\{\beps^\prime : \beps^\prime \prec \beps\}$. The interleaved, two-layered stick-breaking construction is as follows:
\begin{equation}
\label{eqn:tstick}
\begin{aligned}
\nu_\beps \sim \text{Beta}(1,\alpha(|\beps|)) \ ; \quad
\psi_\beps \sim \text{Beta}(1,\gamma) \ ;\quad
\omega_\kappa = \nu_\kappa \ ; \\
\omega_{\beps}  = \nu_\beps \varphi_\eps \prod_{\beps^\prime \prec \beps}\varphi_{\beps^\prime}(1-\nu_{\beps^\prime}) \ ; \quad
\varphi_{\beps\eps_i}  = \psi_{\beps\eps_i} \prod_{j=1}^{\eps_i-1} (1-\psi_{\beps_j}) \ .
\end{aligned}
\end{equation}
The $\nu_\beps$ and $(1-\nu_\beps)$ determine the amount of mass allocated to $\beps$ and its descendants respectively, whereas $\{\varphi_\beps\}$ determines the probability of a particular sequence of children. The construction ensures that the mixing weights $\{\omega_\beps\}$ sum to one. The parameters $\alpha$ and $\gamma$ control the height and the width of the tree respectively. Note that the concentration parameter $\alpha(\cdot)$ is a function of the depth of the tree ($\alpha(\cdot):\N \to \R^+$) and is defined to be $\alpha(j)=\lambda^j \alpha_0$ with $\alpha_0 >0$ and $\lambda \in (0,1]$ \cite{AdamsGJ10}. 

\subsection*{PhyloSub model}
We follow Shah \emph{et al.} \cite{RothS12,Shah12} to model the allelic count data. For each genetic variant that is detected by high-throughput sequencing methods, cells containing the genetic variant is referred to as variant population and those without the variant as reference population. Let $\Sigma = \{A,C,G,T\}$ denote the set of nucleotides. Let $a_i$ and $b_i$ denote the number of reads matching the reference allele $\textrm{A} \in \Sigma$ and the variant allele $\textrm{B} \in \Sigma$ respectively at position $i$, and let $d_i = a_i+b_i$. The genotype $g \in \{\textrm{A}, \textrm{B}, \textrm{AA}, \textrm{AB}, \textrm{BB}, \textrm{AAA}, \ldots\}$ would depend on the copy number at the variant location. Let $\mu_i^r \in \{\mu_i^\textrm{A},\mu_i^\textrm{AA},\mu_i^{\textrm{AAA}},\ldots\}$ denote the probability of sampling a reference allele from the reference population. This value depends on the error rate of the sequencer. Let $\bm{\mu}_i^v$ denote a vector whose entries, $\mu_i^{v:g} \in \{\mu_i^\textrm{B},\mu_i^\textrm{AB},\mu_i^{\textrm{BB}},\ldots\}$, are the probabilities of sampling a reference allele from the variant population with genotype $g$ at position $i$. Let $\bm{\pi}_i$ denote the vector whose entries, $\pi_i^g \in \{\pi^\textrm{B}_i,\pi_i^\textrm{AB},\pi_i^{\textrm{BB}},\ldots\}$, are the probabilities of the variant population at position $i$ to have the genotype $g$. Let $\bm{\delta}_i$ denote the pseudo-count parameters of the Dirichlet prior over $\bm{\pi}_i$. Let $G_i$ denote the genotype of the variant population at position $i$. Let $\tilde{\phi}_i$ denote the fraction of cells from the variant population, i.e., the SNV population frequency at position $i$, and $1-\tilde{\phi}_i$ denote the fraction of cells from the reference population at position $i$. The observation model for allelic counts has the following generative process \cite{Shah12}:
\begin{equation}\label{eqn:pm}
\begin{aligned}
& \mathcal{G}  \sim  \text{DP}(\alpha,H) \ ; \quad
\tilde{\phi}_{i}  \sim \mathcal{G} \ ; \quad
\bm{\pi}_i   \sim \text{Dirichlet}(\bm{\delta}_i) \ ; \quad
G_i  \sim \text{Categorical}(\bm{\pi}_i) \ ;   \\
& a_i \mid d_i,G_i = g,\tilde{\phi}_i,\mu^r_i,\bm{\mu}^{v}_i  \sim \text{Binomial}(d_i,(1-\tilde{\phi}_i)\mu^r_i+\tilde{\phi}_i \mu_i^{v:g}) \ .
\end{aligned}
\end{equation}
The posterior distribution of $\tilde{\phi}_i$ is
\[
p(\tilde{\phi}_i \mid a_i,d_i,\mu_i^r,\bm{\mu}_i^{v},\bm{\pi}_i,\bm{\delta}_i) \propto \sum_g p(a_i \mid d_i,G_i = g,\tilde{\phi}_i,\mu_i^r,\bm{\mu}_i^{v}) p(G_i \mid \bm{\pi}_i)  p(\bm{\pi}_i \mid \bm{\delta}_i) p(\tilde{\phi}_i) \ .
\]
Each of the terms appearing inside the summation over genotypes is the probability distribution of a Dirichlet compound multinomial (with a single draw) \cite{Minka12}. The posterior distribution can thus be rewritten as
\begin{equation}
\label{eqn:posterior}
p(\tilde{\phi}_i \mid \cdot) \propto \sum_g \left[ \frac{\prod_{g^\prime \neq g} \mathrm{\Gamma}(\delta_i^{g^\prime}) \times \mathrm{\Gamma}(\delta_i^g +1)}{\mathrm{\Gamma}(\sum_{g^\prime} \delta_i^{g^\prime}+1)} \right] \text{Binomial}(a_i; d_i,(1-\tilde{\phi}_i)\mu_i^r+\phi_i \mu_i^{v:g}) p(\tilde{\phi}_i) \ ,
\end{equation}
where $\mathrm{\Gamma}(\cdot)$ is the Gamma function.

\begin{figure}[!t]
\centering
\includegraphics{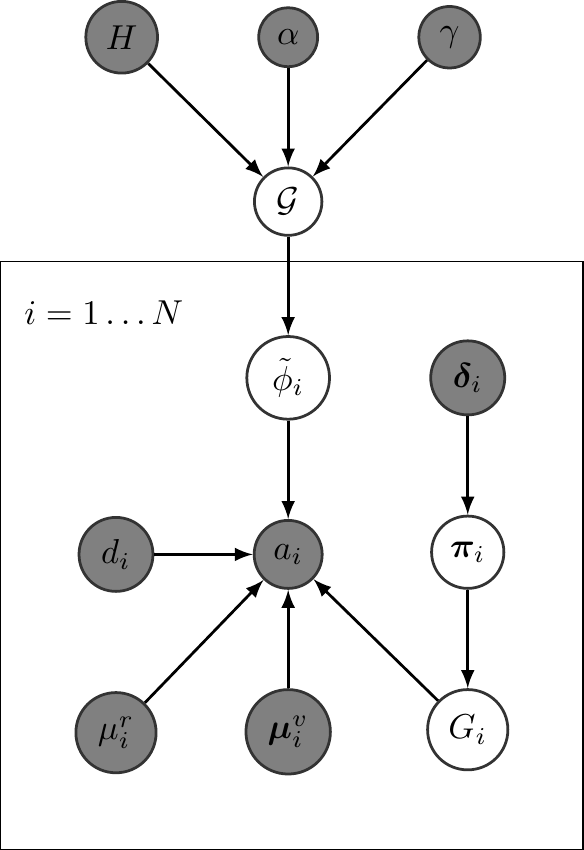}
\caption{{\bf PhyloSub graphical model for single sample.} Probabilistic graphical model for allelic counts with tree-structured stick-breaking process prior. Observed variables and hyperparameters (inputs to the model) are indicated in shaded nodes.}
\label{fig:plate}
\end{figure}

The Dirichlet process prior $\text{DP}(\alpha,H)$ in the observation model of allelic counts \eq{eqn:pm} is useful to infer groups of mutations that occur at the same SNV population frequency \cite{Shah12}. Furthermore, being a nonparametric prior, it is useful to avoid the problem of selecting the number of groups of mutations \emph{a priori}. However, it cannot be used to model the clonal evolutionary structure which takes the form of a rooted tree. In order to model this, we propose to use the tree-structured stick-breaking process prior (\ref{eqn:tstick}) described in the previous section. 
The probabilistic graphical model for allelic counts with the tree-structured stick-breaking process prior is shown in Figure \ref{fig:plate}. 
Inputs to the model including the hyperparameters are indicated in shaded nodes, whereas the latent variables including the set of SNV population frequencies $\{\tilde{\phi}_i\}$ are indicated in unshaded nodes. The prior/base distribution $H$ of the SNV population frequencies is the uniform distribution $\text{Uniform}(0,1)$ for the root node and $\text{Uniform}(0,\phi_{\text{par}(v)}-\sum_{w \in \mathcal{S}(v)} \phi_w)$ for any other node $v$ in the tree, where $\text{par}(v)$ denotes the parent node of $v$ and $\mathcal{S}(v)$ is the set of siblings of $v$. This ensures that the clonal evolutionary constraints (discussed in the next section) are satisfied when adding a new node in the tree. The crucial difference between our model and the model of Shah \emph{et al.} \cite{Shah12} is that we use a tree-structured stick-breaking process instead of a Dirichlet process (cf. \eq{eqn:pm}) to generate the set of SNV population frequencies $\{\tilde{\phi}_i\}$. Given this model and a set of $N$ observations/inputs $\{(a_i,d_i,\mu_i^r,\bm{\mu}_i^v,\bm{\delta}_i)\}_{i=1}^N$, the tree structure and the SNV population frequencies $\{\tilde{\phi}_i\}$ are inferred using Markov chain Monte Carlo sampling. In particular, we use Gibbs sampling \cite{AdamsGJ10} to generate posterior samples of the SNV population frequencies \eq{eqn:posterior}. Each iteration of the Gibbs sampler involves multiple subsampling procedures: sampling cluster assignments  $\{z_i\}$, sampling stick lengths $\nu_{\beps}$ and $\psi_{\beps \eps_i}$, sampling stick-breaking hyperparameters  $\alpha_0$, $\gamma$ and $\lambda$, and sampling the SNV population frequencies  $\{\tilde{\phi}_i\}$. Our main algorithmic contribution, described below, is a method to sample SNV population frequencies in such a way that the tumor evolution proceeds according to the assumptions from the clonal evolutionary theory. The rest of the subsampling procedures follow directly from Adams \emph{et al.} \cite{AdamsGJ10} and we refer the reader to it for further technical details.

\subsection*{Sampling SNV population frequencies}
Given the current state of the tree structure, we sample SNV population frequencies in such a way that the SNV population frequency $\phi_v$ of every non-leaf node $v$ in the tree is greater than or equal to the sum of the SNV population frequencies of its children. To enforce this constraint, we introduce a set of auxiliary weights $\{\eta_v\}$, one for each node, that satisfy $\sum_v \eta_v =1$, and rewrite the observation model for allelic counts \eq{eqn:pm} explicitly in terms of these weights resulting in the following posterior distribution:
\begin{equation}
\label{eqn:posterior1}
p(\tilde{\eta}_i \mid a_i,d_i,\mu_i^r,\bm{\mu}_i^{v},\bm{\pi}_i,\bm{\delta}_i) \propto \sum_g p(a_i \mid d_i,G_i = g,\tilde{\eta_i},\mu_i^r,\bm{\mu}_i^{v}) p(G_i \mid \bm{\pi}_i)  p(\bm{\pi}_i \mid \bm{\delta}_i) p(\tilde{\eta}_i) \ ,
\end{equation}
where we have used $\{\tilde{\eta}_i\}$ to denote the auxiliary weights for each SNV. The prior/base distribution of the auxiliary weights is defined such that it is 1 for the singleton root node and $\text{Uniform}(0,\eta_{\text{par}(v)})$ for any other node $v$ in the tree, where $\text{par}(v)$ denotes the parent node of $v$. When a new node $w$ is added to the tree, we sample $\eta_w \sim  \text{Uniform}(0,\eta_{\text{par}(w)})$ and update $\eta_{\text{par}(w)} \gets \eta_{\text{par}(w)} - \eta_w$. This ensures that $\sum_v \eta_v =1$. 

This change is crucial as it allows us to design a Markov chain that converges to the stationary distribution of  $\{\eta_v\}$. The SNV population frequency for any node $v$ can then be computed via 
\begin{equation}
\label{eqn:eta2phi}
\phi_v= \eta_v + \sum_{w \in \mathcal{D}(v)} \eta_w = \eta_v + \sum_{w \in \mathcal{C}(v)} \phi_w \ ,
\end{equation}
where $\mathcal{D}(v)$ and $\mathcal{C}(v)$ are the sets of all descendants and children of node $v$ respectively. This construction ensures that the SNV population frequencies of mutations appearing at the parent node is greater than or equal to the sum of the frequencies of all its children. The procedure to generate a random sample of SNV population frequencies is given in Algorithm \ref{alg:cons_params} where we 
generate $(\eta_v,\phi_v)$ for every node $v$ by traversing the tree in a breadth-first fashion. The input to this algorithm is the current state of the tree $\mathcal{T} = (V,E)$ where $V$ is the set of vertices and $E$ is the set of edges, and the output is a multi-dimensional sample of SNV population frequencies $\bm{\phi}=\{\phi_1,\phi_2,\ldots,\phi_{|V|}\}$ (where $|V|=K$) and the corresponding auxiliary weights $\bm{\eta}=\{\eta_1,\eta_2,\ldots,\eta_{|V|}\}$. A sample from this algorithm is shown in Figure \ref{fig:cons_params}.

\begin{figure}[ttt!]
\begin{algorithm}[H]
\caption{Algorithm to generate SNV population frequencies satisfying the assumptions from clonal evolutionary theory.}
\label{alg:cons_params}
\renewcommand{\algorithmicrequire}{\textbf{Input:}}
\renewcommand{\algorithmicensure}{\textbf{Output:}}
\begin{algorithmic}[1]
	\REQUIRE Rooted tree $\mathcal{T} = (V,E)$ with root node $r$
	\ENSURE $\bm{\eta}=\{\eta_1,\eta_2,\ldots,\eta_{|V|}\}$, $\bm{\phi}=\{\phi_1,\phi_2,\ldots,\phi_{|V|}\}$
	\medskip
	\STATE create a queue $Q$
	\STATE Q.enqueue($r$)
	\WHILE{Q is not empty}
		\STATE v = Q.dequeue()
		\IF{v is root}
			\STATE $\phi_v = 1$
			\STATE $s_v \sim \text{Uniform}(0,1)$ 
			\STATE $\eta_v = \phi_v \cdot s_v$
		\ENDIF	
		\STATE $m_v = \phi_v-\eta_v$ \COMMENT{mass assigned to children of $v$}
		\FOR{c in children of v}
			\STATE $r_c \sim \text{Uniform}(0,1)$ \COMMENT{distribute mass}
		\ENDFOR
		\STATE $\bm{r} = m _v\cdot \bm{r}/\sum_c r_c$		
		\FOR{c in children of v}
			\STATE $\phi_c = r_c$
			\STATE $s_c \sim \text{Uniform}(0,1)$ 
			\STATE $\eta_c=\phi_c \cdot s_c$
			\STATE Q.enqueue($c$)
		\ENDFOR
	\ENDWHILE
	\RETURN $\bm{\eta}$, $\bm{\phi}$	
\end{algorithmic}
\end{algorithm}
\end{figure}

\begin{figure}[!t]
\centering
\includegraphics{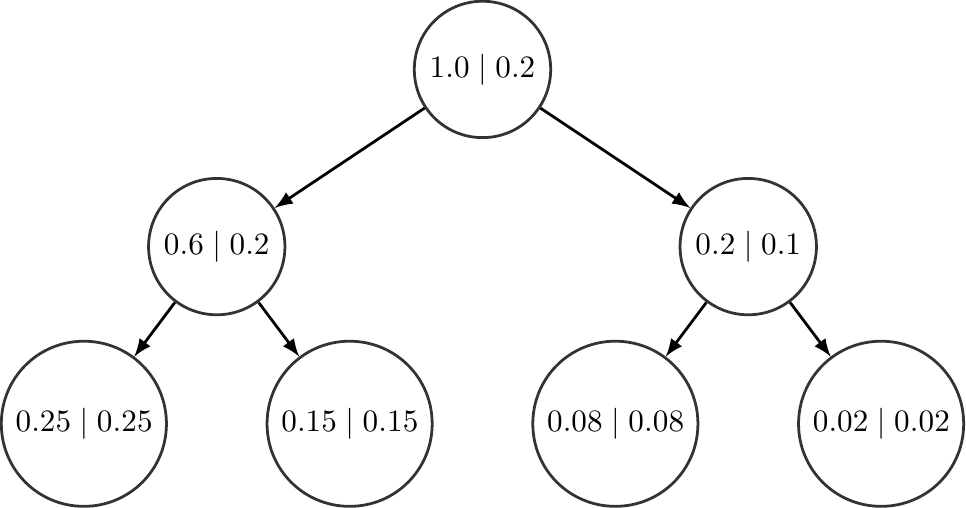}
\caption{{\bf Example of SNV population frequencies generated using Algorithm \ref{alg:cons_params}}. The labels of the nodes are its corresponding SNV population frequencies and weights ($\phi \mid \eta$). Note that $\sum_{v \in V} \eta_v =1$ and  $\phi_v \geq \sum_{w \in \mathcal{C}(v)} \phi_w$ for every non-leaf node $v$.}
\label{fig:cons_params}
\end{figure}

We use Metropolis-Hastings algorithm to sample from the posterior distribution of the auxiliary weights $\{\tilde{\eta_i}\}$ \eq{eqn:posterior1} as shown in Algorithm \ref{alg:mh_params} and derive the SNV population frequencies from these samples. We use an asymmetric Dirichlet distribution as the proposal distribution. This ensures that the Markov chain converges to the stationary distribution of $\{\tilde{\eta_i}\}$. The inputs to the sampling algorithm are the current state of the tree $\mathcal{T} = (V,E)$, a scaling factor $\sigma$, and the number of iterations $T$. The output is a sample from the posterior distribution of $\bm{\eta}=\{\eta_1,\eta_2,\ldots,\eta_{|V|}\}$ and its corresponding $\bm{\phi}=\{\phi_1,\phi_2,\ldots,\phi_{|V|}\}$. 

\begin{figure}[ttt!]
\begin{algorithm}[H]
\caption{Metropolis-Hastings algorithm to sample from the posterior distribution of the auxiliary weights $\{\eta_v\}$ and compute the SNV population frequencies $\{\phi_v\}$.}
\label{alg:mh_params}
\renewcommand{\algorithmicrequire}{\textbf{Input:}}
\renewcommand{\algorithmicensure}{\textbf{Output:}}
\begin{algorithmic}[1]
	\REQUIRE Rooted tree $\mathcal{T} = (V,E)$, $\sigma$, $T$
	\ENSURE  $\bm{\eta}=\{\eta_1,\eta_2,\ldots,\eta_{|V|}\}$, $\bm{\phi}=\{\phi_1,\phi_2,\ldots,\phi_{|V|}\}$
	\medskip
	\STATE Initialize $\bm{\eta}^{(0)}$ using Algorithm \ref{alg:cons_params}
	\FOR{$t = 1 : T$}
		\STATE /\//draw a proposal state from the Dirichlet distribution with density function $Q(\cdot)$
		\STATE $\bm{\eta}^\prime \sim \text{Dirichlet}(\sigma \bm{\eta}^{(t-1)}+1)$
		\STATE // accept/reject state
		\STATE $a=\log p(\bm{\eta}^\prime \mid \cdot) - \log p(\bm{\eta}^{(t-1)} \mid \cdot) + \log Q(\bm{\eta}^{(t-1)}; \bm{\eta}^\prime, \sigma) - \log Q(\bm{\eta}^\prime; \bm{\eta}^{(t-1)},\sigma)$ 
		\STATE $r \sim \text{Uniform}(0,1)$
		\IF{$\log(r)<a$}
			\STATE $\bm{\eta}^{(t)} \gets \bm{\eta}^{\prime}$
		\ELSE
			\STATE $\bm{\eta}^{(t)} \gets \bm{\eta}^{(t-1)}$
		\ENDIF
	\ENDFOR
	\STATE Compute $\bm{\phi}$ from $\bm{\eta}$ \eq{eqn:eta2phi}
	\RETURN $\bm{\eta}^{(T)}, \bm{\phi}^{(T)}$
\end{algorithmic}
\end{algorithm}
\end{figure}

\subsection*{Extension to multiple tumor samples}
\begin{figure}[!t]
\centering
\includegraphics{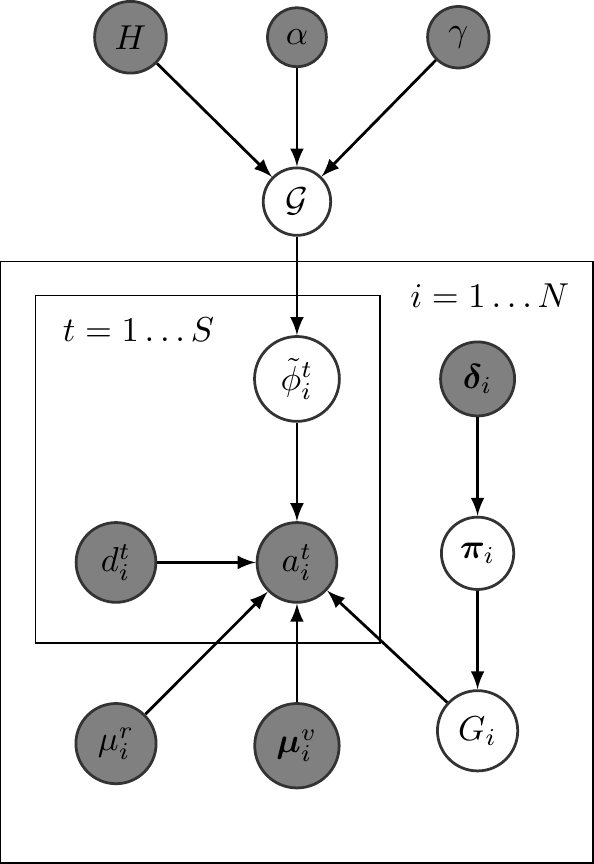}
\caption{{\bf PhyloSub graphical model for multiple samples.} Probabilistic graphical model for allelic counts from multiple samples with a shared tree-structured stick-breaking process prior. Observed variables and hyperparameters (inputs to the model) are indicated in shaded nodes.}
\label{fig:multi_plate}
\end{figure}

PhyloSub (cf. Figure \ref{fig:plate}) can be easily extended for multiple tumor samples. We allow the tree-structured stick-breaking process prior \eq{eqn:tstick} to be shared across all the samples. Let $a^t_i$ and $b^t_i$ denote the number of reads matching the reference and the variant allele respectively at position $i$ for sample $t \in \{1,\ldots,S\}$, and let $d^t_i = a^t_i+b^t_i$. Let $\tilde{\phi}^t_i$ denote the fraction of cells from the variant population, i.e., the SNV population frequency at position $i$ for sample $t$, and $\tilde{\eta}^t_i$ denote its corresponding auxiliary weight. The graphical model of PhyloSub for multiple tumor samples is shown in Figure \ref{fig:multi_plate}.
The main technical difference between the single and the multiple sample models lies in the sampling procedure for SNV population frequencies. In the multiple sample model, we ensure that the clonal evolutionary constraints described in the previous section are satisfied separately for each tumor sample and then make a global Metropolis-Hastings move based on the distribution  $\prod_{t=1}^S p(\bm{\eta}^t \mid \cdot)$, where $\{\bm{\eta}^1,\bm{\eta}^2,\ldots,\bm{\eta}^S\}$ is the set of auxiliary weights for all the tumor samples.

\subsection*{Partial order plot}
We construct a partial order plot to summarize and visualize the trees from all the posterior MCMC samples. It is important to note that the nodes of this partial order plot are the SNVs and not the SNV clusters. The thickness of a directed edge $P \to Q$ in the tree is proportional to the fraction of MCMC samples in which SNV $P$ first appears in a subclonal lineage that was the parent of the subclonal lineage that $Q$ first appears in. The color of the border of the SNVs represents the subclonal lineage cluster that the SNV is placed into posthoc using an algorithm called correlation clustering \cite{BansalBC04}. Note that the main purpose of this clustering algorithm is only to color the nodes in the partial order plot by aggregating the clustering information from all the MCMC samples obtained from our model; this clustering is a summary but does not represent any (possibly quite large) uncertainty in the cluster assignments. The input to this algorithm is a symmetric $N \times N$ co-clustering matrix C, whose elements $C_{ij}$ is the difference between the number of samples in which $i$ and $j$ were assigned to the same SNV cluster and the number of samples in which $i$ and $j$ were assigned to different SNV clusters. The algorithm estimates a symmetric $N \times N$ cluster indicator matrix $Y$, whose elements $Y_{ij} =1$ if $i$ and $j$ are assigned to the same cluster and $Y_{ij}=0$ otherwise. This cluster indicator matrix $Y$ has all the information about the number of SNV clusters as well as the SNVs assigned to each of them.

\subsection*{MCMC settings}
In all the experiments, we fix the number of MCMC iterations to 5,000 with a burn-in of 100 samples. We also fix the number of iterations in the Metropolis-Hastings algorithm to 5,000 and set the scaling factor for the Dirichlet proposal distribution to $\sigma=100$. We run the MCMC samplers multiple times with different random initializations and pick a single run based on the complete-data likelihood trace and its auto-correlation function. We use all the 5,000 samples without thinning \cite{LinkE12} to construct the partial order plots. We use the CODA R package \cite{coda} for MCMC diagnostics to monitor the convergence of the samplers.  The complete-data  log likelihood traces and the corresponding autocorrelation function plots after the burn-in period of 100 samples for all the experiments on AML and CLL datasets are shown in Figures S3 to S7 (Additional file 1). 

\subsection*{Datasets and inputs to PhyloSub}
All datasets used in the experiments,  including details about the inputs to PhlyoSub, i.e., the set of observations $\{(a_i,d_i,\mu_i^r,\bm{\mu}_i^v,\bm{\delta}_i)\}_{i=1}^N$, are provided in the Additional file 2 (Tables S5 -- S10).


\section*{Acknowledgements}
  This work was funded by a National Science and Engineering Research
  Council (NSERC) operating grant and a Early Researcher Award from the Ontario
  Research Fund to QM. WJ and LS were supported by The Ministry of Research and Innovation, Province of Ontario. 

We thank Andrew Roth and Sohrab Shah for sharing a pre-publication version of the PyClone software with us; helping us to install it and to duplicate their published results; and for providing unpublished details of the PyClone observation model in their user manual.

\bibliographystyle{unsrt}
\bibliography{bib}
\end{document}